\documentclass{article}

\usepackage{amsmath,amssymb,graphicx,mlspconf,url}
\usepackage{color}
\usepackage[dvipsnames]{xcolor}

\copyrightnotice{U.S.\ Government work not protected by U.S.\ copyright}

\copyrightnotice{978-1-5090-0746-2/16/\$31.00 {\copyright}2016 Crown}

\copyrightnotice{978-1-5090-0746-2/16/\$31.00 {\copyright}2016 European Union}

\copyrightnotice{978-1-5090-0746-2/16/\$31.00 {\copyright}2016 IEEE}

\toappear{2016 IEEE International Workshop on Machine Learning for Signal Processing, Sept.\ 13--16, 2016, Salerno, Italy}

\title{DYNAMIC MATRIX FACTORIZATION WITH SOCIAL INFLUENCE}
\twoauthors
{Aleksandr Y. Aravkin}
	{Department of Applied Mathematics\\
	University of Washington\\
	Seattle, WA 98195 USA}
  {Kush R. Varshney, and Liu Yang}
	{Mathematical Sciences Department\\
	IBM Thomas J. Watson Research Center\\
	Yorktown Heights, NY 10598 USA}

\newcommand{\R}[1]{{\rm #1}}
\DeclareMathOperator{\tr}{tr}

\begin{document}
\ninept

\maketitle
\begin{abstract}
Matrix factorization is a key component of collaborative filtering-based recommendation systems because it allows us to complete sparse user-by-item ratings matrices under a low-rank assumption that encodes the belief that similar users give similar ratings and that similar items garner similar ratings.  This paradigm has had immeasurable practical success, but it is not the complete story for understanding and inferring the preferences of people.  First, peoples' preferences and their observable manifestations as ratings evolve over time along general patterns of trajectories.  Second, an individual person's preferences evolve over time through influence of their social connections.  In this paper, we develop a unified process model for both types of dynamics within a state space approach, together with an efficient optimization scheme for estimation within that model. The model combines elements from recent developments in dynamic matrix factorization, opinion dynamics and social learning, and trust-based recommendation.  The estimation builds upon recent advances in numerical nonlinear optimization.  Empirical results on a large-scale data set from the Epinions website demonstrate consistent reduction in root mean squared error by consideration of the two types of dynamics.
\end{abstract}

\section{Introduction}
\label{sec:intro}

In many applications, there is a population of learning problems, which we might suppose share an underlying distribution or are otherwise correlated. In the collaborative filtering setting, for instance, each user has a set of items he or she prefers and the fact that one user prefers items A and B increases our confidence that users who prefer item A also tend to prefer item B. Similar types of relatedness also arise in social media contexts. There are now many large and highly-successful online communities, where each user can be modeled as a learning problem (for instance, for selecting advertisements, search results, or restaurant recommendations), and there is significant benefit to be found in the correlations and patterns that extend across users.  However, there is every reason to suspect that the same connections between learning problems that allow better group models in the static case can also help in the dynamic setting as well. For instance, we might suppose that the drifting behaviors of the learning problems are also correlated, and this correlation can be used to model the group's drift, before adjusting to  each individual member.

This perspective has led to the extension of static matrix factorization-based approaches for recommendation to incorporate rich temporal models of the change in preferences over time \cite{SunVS2012,ChuaOL2013,SunPV2014,LoLC2015}.  In these approaches, the rows of the data matrix represent the users and columns the items.  Such approaches use a state space model to capture temporal dynamics, where the state is the set of user factors, and typically use a restrictive class of models to model dynamics (typically linear), and errors ( typically Gaussian), to allow for standard estimation techniques based on the Kalman filter. Scalability is an important issue in all approaches, as both the state space and measurement space can be quite large. 

Two contributions of this paper are (1) relaxing restrictions of previous work on dynamic matrix factorization by allowing a wide variety of nonlinear and non-Gaussian state space models, and (2) showing how 
to design tractable and scalable inference for large-scale problems.  We build on the optimization viewpoint on Kalman smoothing \cite{AravkinBP2014}, stepping away from the forward-backward recursion that is typically used and 
instead formulating a single large optimization problem to be solved using quasi-Newton algorithms.  

The third contribution is to incorporate dynamic phenomena in user behavior: \emph{social influence}, 
which is separate from the general group-level user rating trajectories captured in the temporal models of \cite{SunVS2012,ChuaOL2013,SunPV2014,LoLC2015}.  
A person may eat at a restaurant with a menu that he or she does not much fancy if a group of friends has decided to eat there.  A legislator may vote for a bill sponsored by a second legislator if that second legislator voted for a past bill sponsored by the first legislator.  In general, a person's emotions, preferences, opinions, decisions, and actions are affected by other people.  Social influence includes conformity, compliance, and obedience as various manifestations.  
Opinion dynamics and social learning are models for these types of effects as they evolve over time \cite{AcemogluO2011,LiuWZ2015}.  Social influence in recommendation via opinion dynamics has been considered in \cite{ShangHKC2011}, but not within the matrix factorization paradigm.

We tackle the challenge of modeling social influence in a unified manner with dynamic matrix factorization by incorporating a regularization term for the dynamics that can easily incorporate known social influence structure 
via the graph Laplacian.  In particular, we assume that we are able to observe the existence of social connections among users as a graph at all times.  Such data can be extracted from websites such as Epinions in which users publicly declare which other users they trust.  Given the observed temporally-evolving social influence graph, we include a regularization term that imposes the belief derived from opinion dynamics and social learning theory that future preferences of a user should be similar to the preferences of users that he or she trusts.  Mathematically, this is encoded using the Laplacian of the social influence graph.  Such a term has been used previously in static, but not dynamic, settings to impose similarity \cite{GaoTC2013,HuangNHTL2013}.  We are able to include this term into the overall formulation because of the flexibility afforded to us by the optimization viewpoint on Kalman smoothing; it would not have been able to be included otherwise.

%We tackle the challenge of modeling social influence in a unified manner with dynamic matrix factorization by expanding the process (a.k.a.\ forward) component of the state space model.  In particular, we assume that we are able to observe the existence of social connections among users as a graph at all times.  Such data can be extracted from websites such as Epinions in which users publicly declare which other users they trust.  Given the observed temporally-evolving social influence graph, we include a process term that imposes the belief derived from opinion dynamics and social learning theory that future preferences of a user should be similar to the preferences of users that he or she trusts.  Mathematically, this is encoded using the Laplacian of the social influence graph.  Such a term has been used previously in static, but not dynamic, settings to impose similarity \cite{GaoTC2013,HuangNHTL2013}.  We are able to include this term into the overall formulation because of the flexibility afforded to us by the optimization viewpoint on Kalman smoothing; it would not have been able to be included otherwise.

Our use of the trust graph to inform recommendation shares similarity with recent papers such as \cite{YangLLL2013,FangBZ2014,TangGSBL2015,GuoZY2015}, but those formulations are for the static, not dynamic, setting.  Moreover, the specific way in which the social influence graph affects the objective is different than our formulation.  The evolution of the trust graph over time is analyzed in \cite{TangGSBL2015}, but the temporal insights are not used directly in the recommendation task.

The final contribution herein is an experimental study on real-world large-scale ratings data that shows how including both the general preference trajectory dynamics and the more individualized social influence dynamics to improve predictive accuracy.  In particular, we conduct the study on data from the Epinions website, the only available large, real-world data we know of containing a time-varying trust graph along with the more typical time-varying user-item ratings matrix.  We show an improvement in root mean squared error (RMSE) of dynamic matrix factorization with social influence for a large range of choices of the rank parameter (number of factors) in the matrix factorization.

%The remainder of the paper is organized as follows.  In Section~\ref{sec:background}, we set forth notation and introduce background material on dynamic matrix factorization and the imposition of similarity using a graph Laplacian.  In Section~\ref{sec:formulation}, we introduce our new formulation for dynamic matrix factorization with social influence.  In Section~\ref{sec:optimization}, we discuss how to carry out estimation in the proposed model using an optimization approach.  Section~\ref{sec:empirical} discusses the Epinions data set and presents empirical results.  Section~\ref{sec:conclusion} summarizes the contributions and presents ideas for future research.
\section{Background}
\label{sec:background}

In this section, we review static factorization, 
and show how to extend from static formulations to dynamic matrix factorization. 

\subsection{Notation and Static Matrix Factorization}

Suppose we are interested in the preferences of $m$ users for $n$ products, 
where some users have expressed their preferences for some products, stored in the vector $z \in \mathbb{R}^p$, 
with $1 \leq p \leq mn$. 
%In many applications, $m<<n$, i.e. there are far more users than products; however, both $m$ and $n$ are large. 
Let $R \in \mathbb{R}^{m\times n}$ denote the full matrix listing all preferences;
$R$ is observed only through the dataset $z$:
\[
z = \mathcal{A}(R),
\] 
where $\mathcal{A}$ is an operator from $\mathbb{R}^{m\times n} \rightarrow \mathbb{R}^p$. 

%Low-rank matrix approximation for estimating preferences.
Factorized matrix formulations look for a low rank representation $R = UV^T$, 
where $U \in \mathbb{R}^{m\times k}$, and $V \in \mathbb{R}^{n\times k}$.  
The approach requires the modeler to select the latent dimension $k$, typically 
$k << \min(m,n)$.

The factorized representation allows a fast computation of $\mathcal{A}(R)$.
Note that 
\[
R_{ij} = \langle U_i, V_j \rangle,
\]
an inner product between two vectors of length $k$. Therefore, 
$\mathcal{A}(UV^T)$ can be computed in exactly $kp$ operations, 
which is a key point for tractable approaches in large-scale settings. 

Optimization formulations to obtain factors $U,V$ are of the form 
\begin{equation}
\label{eq:factor}
\min_{U,V} \rho\left(z - \mathcal{A}(UV^T)\right) + \phi_1(U) + \phi_2(V),
\end{equation}
where $\rho$ is a measure of misfit between observed and predicted data
(often least squares), $\phi_1$ and $\phi_2$ are regularization penalties, and
\begin{equation}
\label{eq:sample}
\mathcal{A}(UV^T) = A (V\otimes I)\mathrm{vec}(U),
\end{equation}
with $A \in\mathbb{R}^{p \times mn}$ a sparse mask that 
selects the observed entries and $\mathrm{vec}$ the vectorization operator. 

Problem~\eqref{eq:factor} is nonconvex, but has been tremendously successful 
in practice. Factorization-based approaches allow 
matrix completion for extremely large-scale 
systems by avoiding costly SVD computations~\cite{Srebro2005,Lee2010,RechtRe:2011,aravkin2014fast}.  
%The main idea is to parametrize the matrix $X$ as a product, 
%\begin{equation}\label{product}
%X = LR^T\;,
%\end{equation}
%and to optimize over the factors $L,R$. If $X \in \mathbb{R}^{n\times m}$, then $L \in \mathbb{R}^{n\times k}$, 
%and $R \in \mathbb{R}^{m \times k}$. The decision variable therefore has dimension $k(n +m)$, rather than $nm$; giving tremendous savings when $k \ll m,n$.
%{
%The asymptotic computational complexity of factorization approaches is the same as that of partial SVDs, as both methods are dominated by an O(nmk) cost; 
%the former having to form $R = LR^T$, and the latter computing partial SVDs, at every iteration. However, in practice the former operation is much simpler 
%than the latter, and factorization methods outperform methods based on partial SVDs.  In addition, factorization methods keep an explicit bound on the 
%rank of all iterates, which might otherwise oscillate, increasing the computational burden.  }

While we are interested in dynamic settings, we 
use an approach analogous to~\eqref{eq:factor} to initialize our 
(convex) dynamic inference formulation. 
In the following section, we detail the dynamic formulation, and then 
explain the initialization procedure we use prior to  
entering the dynamic phase.

\subsection{Dynamic Matrix Factorization}
 Datasets that track product preferences have a longitudinal structure, 
 as users continue to evaluate products in time. 
We are most interested in dynamic settings where changes in user preferences can be modeled. 
The symmetry of $U$ and $V$ in~\eqref{eq:factor} is therefore broken; indeed, we are much 
more interested in modeling and inference of {\it user dynamics}, so we focus on $U$.

The general dynamic model is as follows. We assume our dataset 
has a natural representation over $N$ observation times, $t_1, \dots, t_N$. 
We assume that $U_t$ is an unknown time series (total size $m\times k\times N$) to be determined, 
and define some process transition matrix $G_t$, a {\it known linear process} that describes the transition. 
The models $V_t$ can be estimated e.g. by solving~\eqref{eq:factor} independently, 
or obtaining an averaged model $V$ over $N$ time points. 

Treating $U_t$ as the unknown state and $V_t$ as a known measurement model, we arrive at 
the linear model 
\begin{equation}
\label{eq:dynamics}
\begin{aligned}
	U_{t+1} & = G_t U_t + \epsilon_t,\\
	z_t &=\mathcal{A}_t\left(U_{t}V_{t}^T\right) + \nu_t
\end{aligned}
\end{equation}
where $\epsilon_t$ describes process noise,  and $\nu_t \in \mathbb{R}^{p_t}$ is observation noise
with known covariance matrix $S_t$.
Note that dimensions of observation vectors $z_t$ can vary 
between time points, hence $z_t \in \mathbb{R}^{p_t}$. 

One of our main contributions is to define an appropriate model $G_t$ that can capture 
inertia, or smoothness, in user preferences, and combine it with measurement 
information and social trust. Before we discuss the dynamic model, 
we review how information about influence can be brought to bear 
on the inference problem.

%%%%%%%%%%%%%%%%%%%%%%%%%%%%%%%%%%%%%%%%%%%%%%%%%%%%%%%
\subsection{Initialization Procedure}
%%%%%%%%%%%%%%%%%%%%%%%%%%%%%%%%%%%%%%%%%%%%%%%%%%%%%%%

Inference over the dynamic model~\eqref{eq:dynamics} is a convex problem in $\{U_t\}$, 
as long as $\{V_t\}$ are assumed fixed. 
However, to obtain these measurement models $V_t$, we need to initially factor each 
matrix $R_t$ into form $U_tV_t$. The key idea here is to extract $V_t$, which then 
become part of the fixed measurement model, as we track $U_t$.

To obtain the factorization $R_t \approx U_tV_t^T$, we follow the approach of~\cite{aravkin2014fast}, 
and solve the problem 
\begin{equation}
\label{eq:initialization}
\begin{aligned}
\min_{U_t, V_t}  & \frac{1}{2}\left(\|U_t\|_F^2 + \|V_t\|_F^2\right) \\
\mathrm{s.t.} & \|b - \mathcal{A}(U_tV_t^T)\|_2 \leq \sigma. 
\end{aligned}
\end{equation}
using publicly available code. 
Formulation~\eqref{eq:initialization} controls the quality of factorization by means of 
both the rank $k$ of the factors, and the regularizer $\frac{1}{2}\left(\|U_t\|_F^2 + \|V_t\|_F^2\right)$;
it is well known (see e.g. ~\cite{Lee2010,RechtRe:2011}) that 
\[
\|R\|_* = \inf_{R = UV^T} \frac{1}{2}\left(\|U_t\|_F^2 + \|V_t\|_F^2\right).
\]
Therefore, every solution $\overline U_t, \overline V_t$ corresponds to $\hat R_t = \overline U_t \overline V_t^T$ 
with $\|R_t\|_* \leq \frac{1}{2}\left(\|\overline U_t\|_F^2 + \|\overline V_t\|_F^2\right)$, where $(\overline U, \overline V)$ minimize
the Frobenius norm over the set
\[
\{U_t, V_t: \|b - \mathcal{A}(U_tV_t^T)\|_2 \leq \sigma \}.
\]
Thus, even if the rank $k$ is picked to be too large, the formulation~\eqref{eq:initialization} maintains control 
of model complexity through minimizing the functional. We performed our experiments over ranks $k = 5, 10, 15, 20$
(see in Figures \ref{fig:k5}--\ref{fig:k20}), and the static RMSE (obtained from the initial factorization~\eqref{eq:initialization}) 
is slightly increasing in $k$ but does not vary much.

%%%%%%%%%%%%%%%%%%%%%%%%%%%%%%%%%%%%%%%%%%%%%%%%%%%%%
\subsection{Similarity Using the Graph Laplacian }
%%%%%%%%%%%%%%%%%%%%%%%%%%%%%%%%%%%%%%%%%%%%%%%%%%%%%

Our main goal is to incorporate the effect of social trust on changes in preferences. 
In particular, suppose that for each time point, we are given matrices $W_t \in \mathbb{R}^{m\times m}$ 
that encode the trust/influence between users.  
Recall that degree matrices corresponding to $W_t$ are given by diagonal $D_t \in \mathbb{R}^{m\times m}$,
with $d_{ii} = \sum_{i'=1}^{m} w_{ii'}$, while graph Laplacian matrices $L_t$ are defined by $L_t = D_t - W_t$.  

We want to look for solutions $U_t$ 
that are more consistent with relationships encoded by $L_t$, 
which makes it more likely for preferences users with mutual trust to evolve in a mutually consistent
manner, i.e. along level sets of the following functional: 
\begin{equation}
\label{eq:LaplacianReg}
\phi_t(U_t) = \tr\left(U_t^TL_tU_t\right).
\end{equation}

The key missing detail is continuity in preferences; or some prior on the {\it smoothness} of preference changes over time. 
In the next section, we show how to incorporate this notion in order to track dynamic preference systems of form~\eqref{eq:dynamics}. 
%while also incorporating~\eqref{eq:LaplacianReg} to guide the dynamic model.

\section{Proposed Formulation}
\label{sec:formulation}

To capture the {\it smoothness} or {\it inertia} of user preferences over time, 
we use a model common to physical systems. In particular, 
we posit the existence of a {\it velocity} state $\dot U$,
together with a simple integration model to link $U$ and $\dot U$:
\begin{equation}
\label{eq:smoothSignal}
\begin{bmatrix} \dot U \\ U \end{bmatrix}^{t+1} = \begin{bmatrix} I & 0 \\ dt & I \end{bmatrix} 
\begin{bmatrix} \dot U \\ U \end{bmatrix}^{t} + w, \quad w \sim N(0, Q). 
%Q = \begin{bmatrix} dt & dt^2/2\\ dt^2/2 &dt^3/3\end{bmatrix}
\end{equation}
This model is frequently used for smooth systems in dynamic inference~\cite{AravkinBP2014}. 
Pairs of elements  $(\dot U_t(i,j), U_t(i,j))$ are pairwise correlated with known covariance~\cite{Jaz}
\begin{equation}
\label{eq:procCov}
Q = \begin{bmatrix} dt & dt^2/2\\ dt^2/2 &dt^3/3\end{bmatrix}.
\end{equation}

Note that our state is no longer just $U$, and model~\eqref{eq:smoothSignal} doubles the state space. 
However, the dynamics are so simple that this has little effect
on computational complexity when properly implemented. 
We now specify the full dynamic model, starting with the defining relationship between 
the state variable $x_t$ and the dynamic preferences $(\dot U_t, U_t)$: 
\begin{equation}
\label{eq:dynamics2}
\begin{aligned}
x_t & := \mathrm{vec}\left(\begin{matrix} \dot U_t \\ U_t\end{matrix}\right), \\
x_{t+1} &= (G_t+\widetilde L_t)x_t + \epsilon_t, \quad \epsilon_t \sim N(0, Q_k),\\
 z_t  &= H_tx_t + \nu_t, \qquad\qquad \nu_t \sim N(0, \sigma^2 I_{p_t}),\\
 G_t &= \begin{bmatrix} I & 0 \\ \Delta t & I\end{bmatrix}, 
\quad \widetilde L_t = \begin{bmatrix} 0 & 0 \\ 0 & L_t\end{bmatrix},\\
 H_t &=  A(V_t \otimes I) \begin{bmatrix} 0 & I \end{bmatrix}x_t,
\end{aligned}
\end{equation}
where 
\[
Q_t = Q \otimes I_{(m+n)k}
\]
for Q in~\eqref{eq:procCov}, 
%$\widetilde L_t$ incorporates~\eqref{eq:LaplacianReg} into the process model, 
and 
we have used the characterization~\eqref{eq:sample} to explicitly write $H_t$.

\section{Estimation Methodology}
\label{sec:optimization}

%Given a sequence of column vectors $\{ u_k \}$
%and matrices $ \{ T_k \}$ we use the notation
%\[
%\R{vec} ( \{ u_k \} )
%=
%\begin{bmatrix}
%u_1 \\ u_2  \\ \vdots \\ u_N
%\end{bmatrix}
%\; , \;
%\R{diag} ( \{ T_k \} )
%=
%\begin{bmatrix}
%T_1    & 0      & \cdots & 0 \\
%0      & T_2    & \ddots & \vdots \\
%\vdots & \ddots & \ddots & 0 \\
%0      & \cdots & 0      & T_N
%\end{bmatrix} .
%\]
In order to write down the full time series
smoothing problem, we use the following 
definitions: 
\[
\begin{aligned}
\mathcal{Q}       & =  \R{diag} ( \{ Q_t \} )
\\
\mathcal{H}       & = \R{diag} (\{H_t\} )
\\
\mathcal{L}        & = \R{diag} (\{\widetilde L_t\})
\end{aligned}\quad \quad
\begin{aligned}
x       & = \R{vec} ( \{ x_t \} )
\\
w      &  = \R{vec} (\{g_0, 0, \dots, 0\})
\\
z      & = \R{vec} (\{z_1,  z_2, \dots, z_N\})
\end{aligned} 
\]
\[
\begin{aligned}
\mathcal{G}  & = \begin{bmatrix}
    I  & 0      &          &
    \\
    -G_2   & I  & \ddots   &
    \\
        & \ddots &  \ddots  & 0
    \\
        &        &   -G_N  & I
\end{bmatrix}.
%\mathcal{L}  &= \begin{bmatrix}
%    0  & 0      &          &
%    \\
%    0   & \sqrt{\lambda_1} L_1 \otimes I  & \ddots   &
%    \\
%        & \ddots &  \ddots  & 0
%    \\
%        &        &   0  & \sqrt{\lambda_N}L_k \otimes I
%\end{bmatrix}.
\end{aligned}
\]
where $g_0:=g_1(x_0)=G_1x_0$.
Using this notation, the full smoothing problem 
can be written 
\begin{equation}\label{fullLS}
\min_{x} f(x) := \frac{1}{2\sigma^2}\|\mathcal{H}x - z\|^2 + \frac{1}{2}\|\mathcal{G}x -w\|_{\mathcal{Q}^{-1}}^2 
+ \frac{\lambda}{2} x'\mathcal{L}x.
\end{equation}

We have now cast the problem as a very large and 
extremely sparse least 
squares system. 
This formulation incorporates both the inertial information from~\eqref{eq:smoothSignal}
and effect of social inference from~\eqref{eq:LaplacianReg}.
%Note that the combined process model 
%includes both dynamic information about {\it smoothness} 
%of preferences over time (encoded by $\mathcal{G}$)
%and social influence information (encoded by $\mathcal{L}$). 
The parameter $\lambda$ controls the relative influence 
of each modeling component, and its effect can be seen 
in Figures \ref{fig:k5}--\ref{fig:k20}. 

We also want to allow the flexibility to replace 
the process and measurement penalties by more robust variants, 
including the Huber loss and other loss functions~\cite{JMLR:aravkin13a}. 
Modelers may also choose to place simple constraints or regularizers
on the state variable $x$. 
Therefore, rather than focusing on the linear system 
$\nabla f(x) = 0$, we treat~\eqref{fullLS} as a general
optimization problem, focusing our complexity analysis on gradient computation. 
%\begin{equation}
%\label{eq:linearSystem}
%(H^\top S^{-1} H + G^\top Q^{-1} G+ \mathcal{L}^T\mathcal{L}) x =  H^\top R^{-1}z + G^\top Q^{-1}w\;.
%\end{equation}
The complicating factor to any approach is the system $\mathcal{H}$. 
From~\eqref{eq:dynamics}, it clear that forming $\mathcal{H}$ 
explicitly is equivalent to computing
\[
z_t = \mathcal{A}_t(U_tV_t^T)
\]
by first forming $U_tV_t^T$ at each time point and then applying a sparse mask,
which has complexity $O(mnp)$, and is intractable even for a single time point!  
In contrast, as discussed earlier, computing $z_t$
by exploiting structure has complexity $O(pk)$, which can be done very quickly. 
Therefore, we are forced to solve~\eqref{fullLS} 
using only matrix-free methods (i.e. methods using matrix-vector products).

\subsection{Complexity of Gradient Computation}
We can proceed to minimize~\eqref{fullLS} using matrix-free
methods. To simplify the analysis, we assume that 
we will use gradient-based optimization, such as steepest 
descent or L-BFGS~\cite{nocedal99}, and compute the 
complexity of each gradient computation. 

%The gradient of~\eqref{fullLS} is given by 
%\[
%\frac{1}{\sigma}\mathcal{H}^*(\mathcal{H}x - z) + (\mathcal{G} + \lambda\mathcal{L})^*\mathcal{Q}^{-1}\left((\mathcal{G} + \lambda\mathcal{L})x - w\right).
%\]
The gradient of~\eqref{fullLS} is given by 
\[
\frac{1}{\sigma}\mathcal{H}^*(\mathcal{H}x - z) + \mathcal{G}^*\mathcal{Q}^{-1}\left(\mathcal{G}x - w\right) 
+ \lambda \mathcal{L}x.
\]

The system $\mathcal{G}$ is block lower bidiagonal, with identity matrices on the main block, and $-G_t$ blocks on the subdiagonal block. 
Therefore, $\mathcal{G}$ has three nonzero diagonals, and so applying $\mathcal{G}$ or $\mathcal{G}^T$ has cost $O(Nkm)$, where $N$ is the number of time steps, $k$ is the chosen rank of $U$, and $m$ is the number of users. The matrix $\mathcal{Q}$ is block diagonal, and its inverse can be computed and applied in $O(Nmk)$ 
operations. 

Next, $\mathcal{L}$ is block diagonal, containing zeros and sub-blocks $L_t$. 
Each matrix $L_t$ is given by $D_t - W_t$, where $D_t$ is diagonal, and $W_t$ represents influence
information, and is typically sparse. If $q$ is the average number of nonzero entries of the influence matrix $W_t$, 
then applying $\mathcal{L}$ has complexity $O(N(m+q)k)$.

Finally we consider $\mathcal{H}$. The number of measurements can change between time points,
but let $p$ be the average number of measurements. Since each measurement $z_t^i$ can be computed 
from the factorization $U_tV_t^T$ by a single inner product in $O(k)$, we can apply $\mathcal{H}$  and $\mathcal{H}^*$ in $O(Nkp)$ operations. 

The final complexity is then given by $O(Nkm + Nkp + Nk(m+q)) = O(Nk(m+p+q))$. 
In particular, it scales linearly with the number of time steps $N$, and the chosen rank $k$.  
Moreover, the number of products $n$ affects the complexity only though $p$, the observed data points 
(the number of observations should grow with both $m$ and $n$ to get meaningful inference). 
This complexity is reasonable, since $O(Nkm)$ is required simply to update
the decision variable $x$.  Furthermore, it would not change if we replaced the least squares
penalty with another smooth penalty $\rho$; additional steps to evaluate $\nabla \rho$ 
would require $O(Nkm)$ operations for the process and $O(Np)$ for the measurement.  
Similarly, any separable regularizer or constraint on $x$ would require either an 
 $O(Nkm)$ operation or $O(Nkm\log(Nkm))$ evaluation to complete.

\section{Empirical Results}
\label{sec:empirical}

In this section, we demonstrate the value of the proposed model by using it to estimate ratings in a data set from the Epinions website.  

\subsection{Description of Data Set}

Epinions was a general consumer review website launched in 1999 and active until March 2014, whose rating and reviewing content was compared favorably to \emph{Consumer Reports}.  Users entered numerical ratings on a one to five scale and text reviews of products and services across a large number of categories, and importantly for this work, the site included community features through which users could indicate which other users they trusted.  %The site also had a reputation system in which reviews were given helpfulness scores, but we do not use that data in this work.

The numerical ratings for all users and items along with the day of posting were scraped as part of the work of \cite{TangGSBL2015}.  Trust and trusted relationships were also scraped along with the day the relationship was established.  Parsed versions of the raw data and the raw data itself are made available by the authors.\footnote{Epinions data is available at \url{http://www.jiliang.xyz/trust.html}.}  Unfortunately the parsed data does not contain a dictionary of user ids that would allow us to connect the ratings for a user with his or her trust relationships, and thus we reparsed the raw data.  We limited ourselves to users with more than ten ratings.

Through our parsing, we obtained ratings data from July 5, 1999 to May 9, 2011 constituting $m = 22164$ users, $n = 305301$ products, and $\sum p_t = 975449$ ratings.  We also obtained trust relationship data from January 17, 2001 to April 19, 2011 constituting 264022 undirected edges created.  We quantized the times into $N = 11$ bins using the same cutoff values as \cite{TangGSBL2015}.  The data is extremely sparse.  

\begin{figure}
\begin{center}
\begin{tabular}{cc}
\includegraphics[width=0.46\columnwidth]{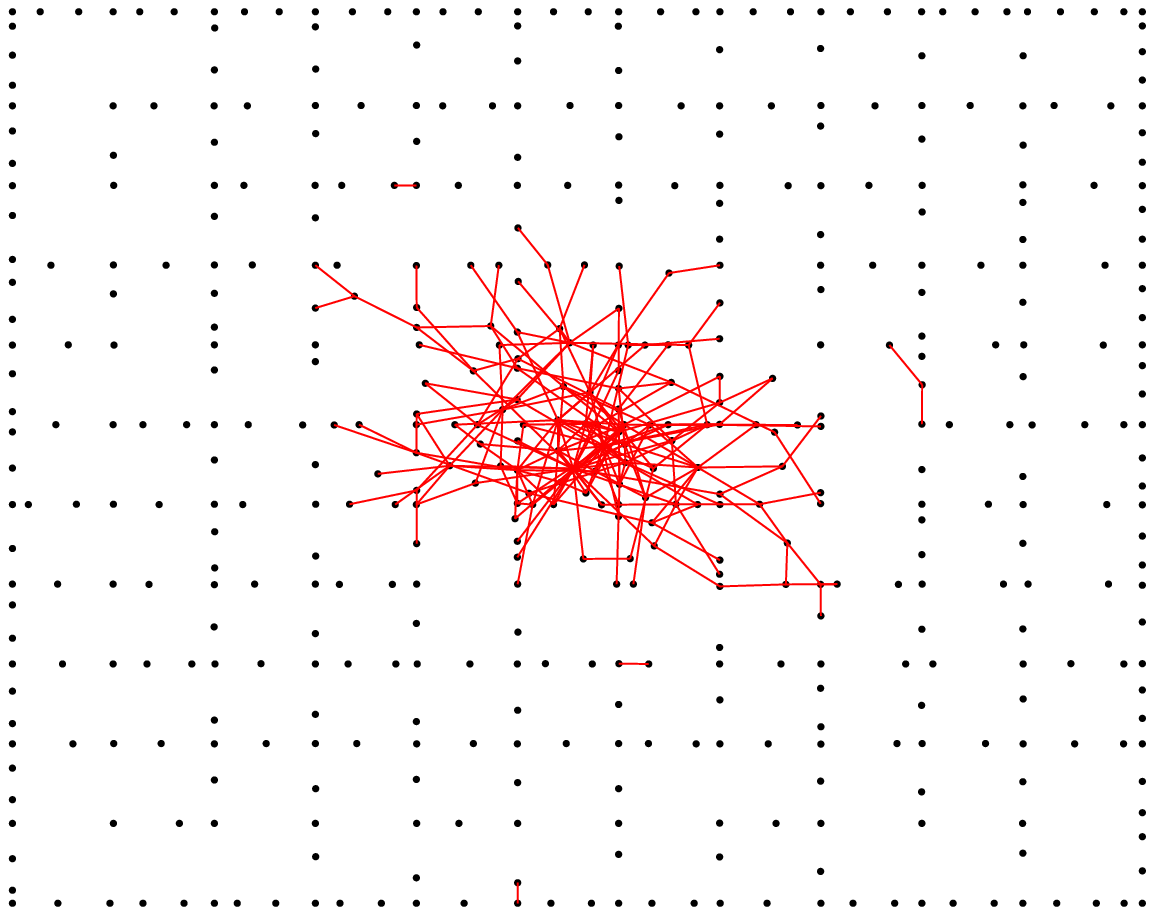} & \includegraphics[width=0.46\columnwidth]{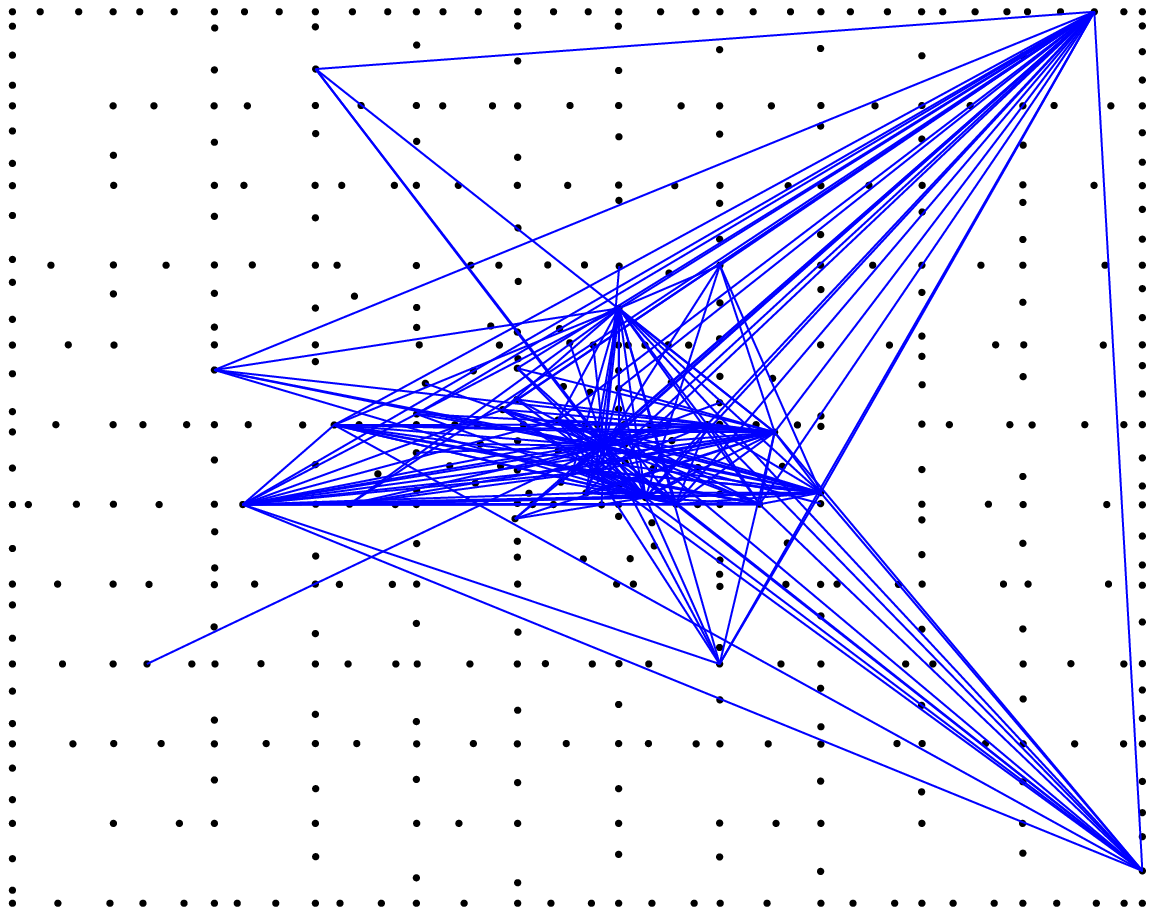} \\
(a) & (b)
\end{tabular}
\caption{(a) Trust relationships and (b) rating similarity above a threshold among 500 random users at $t = 2$.}
\label{fig:edges}
\end{center}
%\vskip -0.2in
\end{figure}

We plot the trust graph for 500 random users on one of the time steps in Figure~\ref{fig:edges}(a).  (If we plot more users, then it is hard to see anything.)  There is some structure to the graph, but most users are not connected.  For comparison, in Figure~\ref{fig:edges}(b), we plot a graph containing edges between users that have similar ratings.  Specifically, this similarity is computed from inner products of completed matrix ratings with $k=5$ and static matrix factorization on the entire set of $m = 22164$ users (without any training/testing split).  A key thing to notice is that the sets of edges do not intersect much, showing that the trust relationships and the ratings do in fact provide complementary information.  %The intersection is shown in Figure~\ref{fig:common_edges}.  This shows that the trust relationships and the ratings do in fact provide complementary information.

\subsection{Experimental Setup and Results}

We split the ratings data randomly within each of the 11 time steps into 50\% training and 50\% testing.  This split is maintained across all experimental settings.  We compare three different models of increasing expressibility: static matrix factorization independently for each of the time steps, dynamic matrix factorization using Kalman smoothing, and dynamic matrix factorization with social influence.  We learn the factors using the estimation procedure described in Section~\ref{sec:optimization} on the training set and then multiply the learned factors out to complete the matrices.  We calculate the average RMSE on the test set, weighted by $p_t$.  Both of the dynamic matrix factorizations require a known $V_t$ for each time; we use the $V_t$ obtained from the static matrix factorizations.  We examine the performance at four different values of the number of latent factors: $k = 5$, $10$, $15$, and $20$.  In the dynamic matrix factorization with social influence, we consider relative weight between the smooth trajectory term and the social influence term across several orders of magnitude: $\lambda = 10^{-5}$, $10^{-4}$, $10^{-3}$, $0.01$, $0.1$, and $1$.

\begin{figure}[t]
\begin{center}
\centerline{\includegraphics[width=0.92\columnwidth]{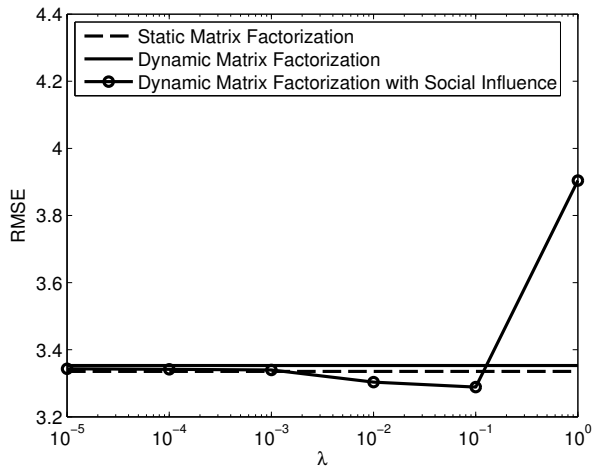}}
\caption{Test accuracy for $k = 5$ factors.}
\label{fig:k5}
\end{center}
\end{figure}

\begin{figure}
\begin{center}
\centerline{\includegraphics[width=0.92\columnwidth]{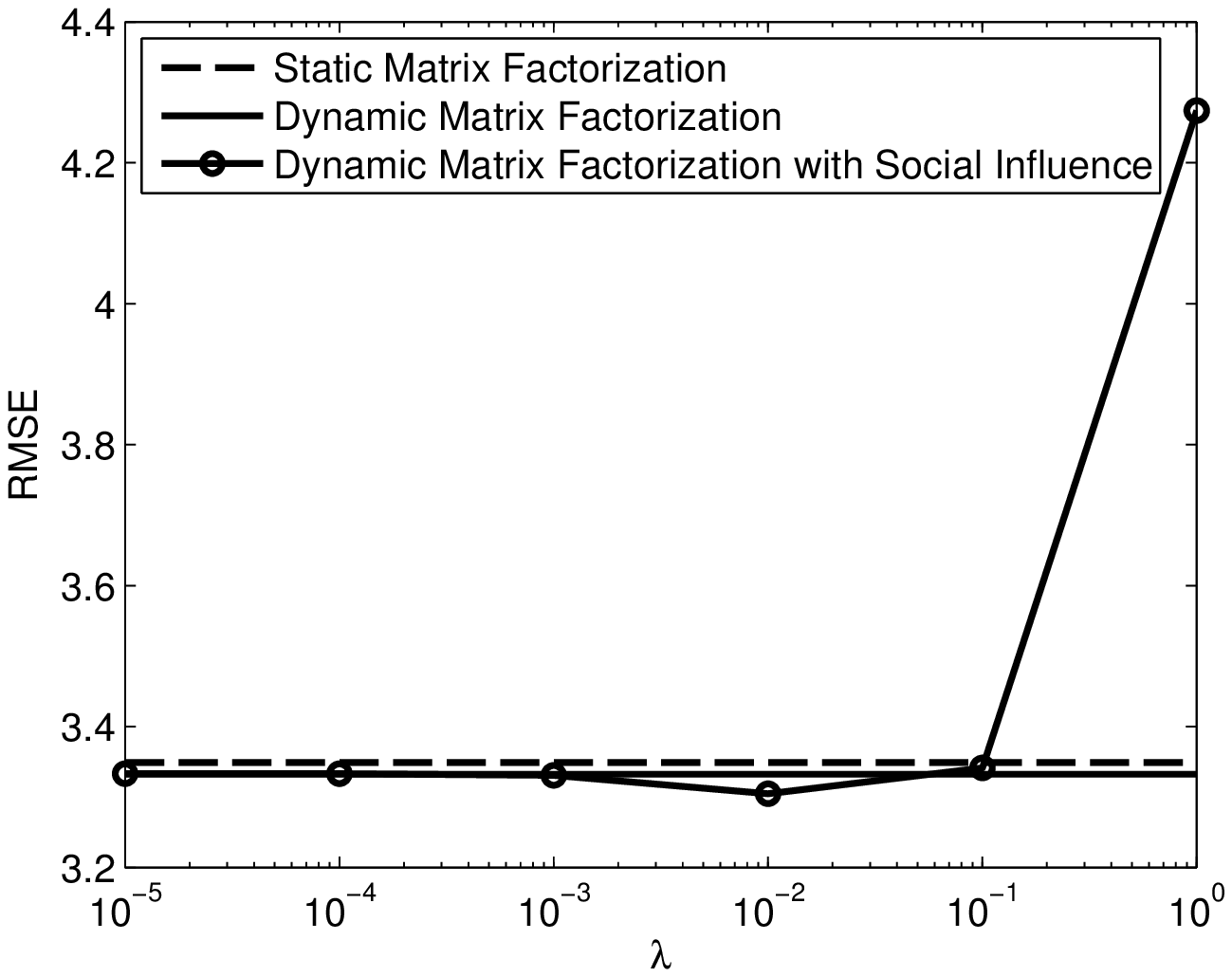}}
\caption{Test accuracy for $k = 10$ factors.}
\label{fig:k10}
\end{center}
\end{figure}

\begin{figure}[t]
\begin{center}
\centerline{\includegraphics[width=0.92\columnwidth]{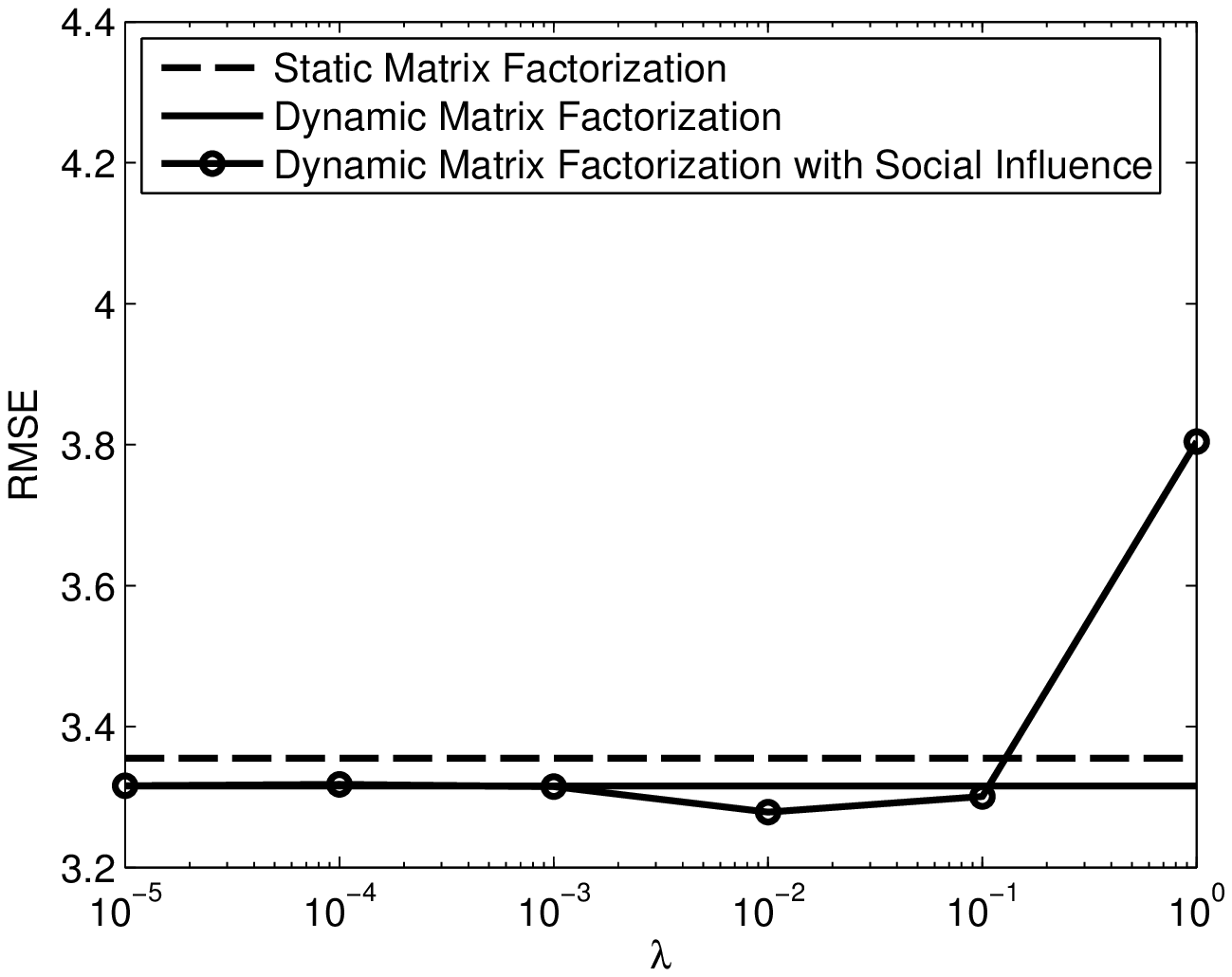}}
\caption{Test accuracy for $k = 15$ factors.}
\label{fig:k15}
\end{center}
\end{figure}

\begin{figure}
\begin{center}
\centerline{\includegraphics[width=0.92\columnwidth]{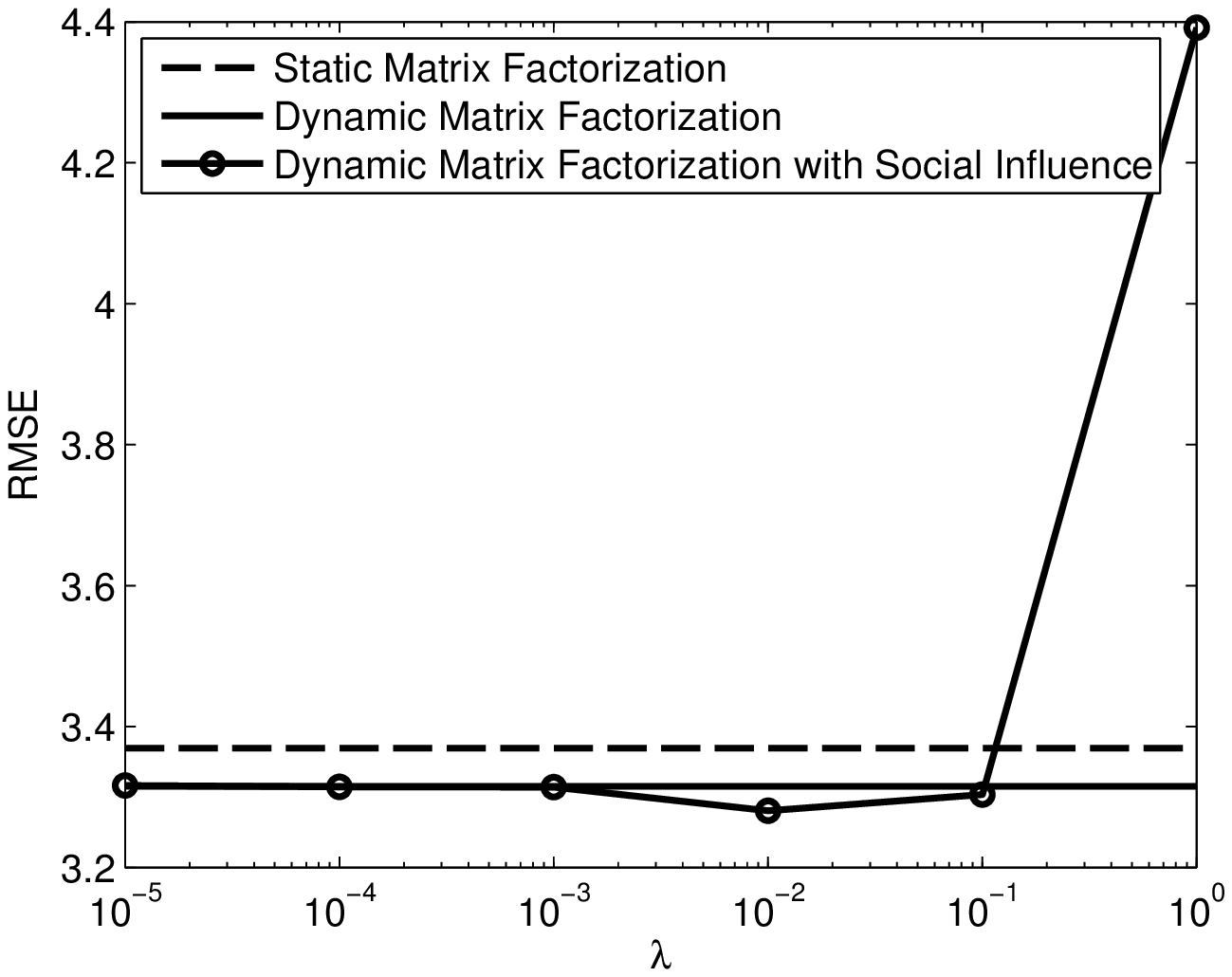}}
\caption{Test accuracy for $k = 20$ factors.}
\label{fig:k20}
\end{center}
\end{figure}

The results are plotted in Figures \ref{fig:k5}--\ref{fig:k20} with each figure giving results for a different value of $k$.  The plots are given as a function of $\lambda$, with the static matrix factorization and dyamic matrix factorization being constant because the respective models do not contain $\lambda$.  The first thing to notice is that except for $k=5$, the error of dynamic matrix factorization is smaller than the error of static matrix factorization.  This behavior recapitulates existing work on dynamic matrix factorization on a large-scale data set.  At $k=5$, the number of factors is so small that the dynamic model is unable to really express itself.  As $k$ increases, the error of the static model increases whereas the error of the dynamic model continues to decrease.  While the improvement in RMSE of the dynamic model over the static model may seem small at first glance, as discussed by \cite{TangGSBL2015}, improvements of the order of magnitude we see are in fact quite valuable.

Now let us turn to the model with the social influence component.  We see that the performance behavior as a function of $\lambda$ is as expected from the structural risk minimization principle.  There is large error caused by overfitting at the very large value of $\lambda = 1$ and an optimal performance at an intermediate value of $\lambda$ around $0.01$, perhaps a bit larger for $k=5$.  In practical operation, $\lambda$ could be determined by cross-validation.  Models with small values of $\lambda$ perform like the dynamic matrix factorization without social influence as they should.  The main thing to notice is that for appropriate choices of $\lambda$, inclusion of the social influence term noticeably improves the performance of ratings prediction.  This is true across a wide range of choices for $k$ and $\lambda$.  Among the values we tested on, the best RMSE was 3.2783 for $k = 15$ and $\lambda = 0.01$, whereas the best RMSE for the static model for $k = 5$ was only 3.3352.  From this analysis, it is apparent that the evolution of people's preferences really is caused by two different phenomena, one general and one individual based on social learning.

% Note use of \abovespace and \belowspace to get reasonable spacing 
% above and below tabular lines. 

%\begin{table}[t]
%\caption{Classification accuracies for naive Bayes and flexible 
%Bayes on various data sets.}
%\label{sample-table}
%\vskip 0.15in
%\begin{center}
%\begin{small}
%\begin{sc}
%\begin{tabular}{lcccr}
%\hline
%\abovespace\belowspace
%Data set & Naive & Flexible & Better? \\
%\hline
%\abovespace
%Breast    & 95.9$\pm$ 0.2& 96.7$\pm$ 0.2& $\surd$ \\
%Cleveland & 83.3$\pm$ 0.6& 80.0$\pm$ 0.6& $\times$\\
%Glass2    & 61.9$\pm$ 1.4& 83.8$\pm$ 0.7& $\surd$ \\
%Credit    & 74.8$\pm$ 0.5& 78.3$\pm$ 0.6&         \\
%Horse     & 73.3$\pm$ 0.9& 69.7$\pm$ 1.0& $\times$\\
%Meta      & 67.1$\pm$ 0.6& 76.5$\pm$ 0.5& $\surd$ \\
%Pima      & 75.1$\pm$ 0.6& 73.9$\pm$ 0.5&         \\
%\belowspace
%Vehicle   & 44.9$\pm$ 0.6& 61.5$\pm$ 0.4& $\surd$ \\
%\hline
%\end{tabular}
%\end{sc}
%\end{small}
%\end{center}
%\vskip -0.1in
%\end{table}

\section{Conclusion}
\label{sec:conclusion}

%Allowing both the user factors and the item factors to evolve over time \cite{AravkinVM2013}. Second state space model in which both the user factor and the item factor evolve over time.  Both $U$ and $V$ together are the state.
%\begin{align}
%	\begin{bmatrix}U^T(t) & V^T(t)\end{bmatrix} &= A(t)\begin{bmatrix}U^T(t-1) & V^T(t-1)\end{bmatrix} + \Phi(t) \\
%	Y^T(t) &= I_{\Omega(t)}\odot \left(V(t)U^T(t) + \Psi(t)\right)
%\end{align}
%
%For the fully nonlinear case, we would have the following process model: 
%\[
%\begin{bmatrix} \dot U \\ \dot V \\ U\\ V \end{bmatrix}^{k+1} = \begin{bmatrix} I & 0 & 0 & 0 \\ 0 & I & 0 & 0 \\ dt & 0 & I & 0 \\ 0 & dt & 0 & I\end{bmatrix} 
%\begin{bmatrix} \dot U \\ \dot V \\ U \\ V \end{bmatrix}^{k} + w, \quad w \sim N(0, Q), \quad 
%Q = \begin{bmatrix} dt & 0 & dt^2/2 & 0 \\ 0 & dt & 0 & dt^2/2 \\ dt^2/2 & 0 & dt^3/3 & 0 \\ 0 & dt^2/2 & 0 & dt^3/3\end{bmatrix}
%\]

In this paper, we considered dynamic modeling of user preferences for products, and 
developed a framework that incorporates trust relationships into these dynamics. The framework
allows the modeler to combine observed preference ratings with three salient features of preference dynamics:
1. Low rank structure: evolving preferences are described by hidden latent states; 2. Time continuity: latent variables controlling preferences change smoothly in time; 3. Trust between users: users influence each other through social relationships. The approach is initialized by obtaining estimates of latent states for each time point where 
data are available, and then is cast as a convex dynamic smoothing problem over the 
observed period. 

For large-scale data, computational complexity becomes a key consideration. Maintaining 
or explicitly forming any structure in the user-product space is prohibitively expensive. 
We cast the inference problem as a structured objective formulation, 
with moderate complexity of $O(N k(m + p + q))$ operations per gradient evaluation,
where $N$ is the number of time periods observed, $k$ is the chosen rank of the 
latent variables, $m$ is the number of users, $p$ is the average number of preference 
observations per time point, and $q$ is the average number of edges in the trust graph. 

Numerical experiments showed that incorporating influence graph information 
into the process model can yield scientifically significant improvements in RMSE. 
The approach requires tuning a trade-off parameter $\lambda$, that controls 
the balance between continuity of preferences across time and trust-graph relationships.

In this work, the social influence or trust relationships are observed and available as a graph at each time step.  One direction for future work is to pose a model and estimation procedure in which we simultaneously infer the social influence graph as part of the state and include a forward model of its time evolution based on theories of opinion dynamics.  This direction of research relates to the social radar method of \cite{WaiSL2015} and will have to deal with issues of identifiability.  Moreover, $\lambda$ can be made time-varying and a part of the state to be estimated as well.  

A significant model enhancement is to consider the case of both $U_t$ and $V_t$ being parts of the state having smooth forward trajectories to be estimated, rather than only $U_t$ being estimated with $V_t$ fixed in advance.  Such an option introduces a highly nonlinear measurement model that cannot be handled by any typical Kalman smoothing approach, but can be handled by the optimization approach \cite{AravkinVM2015}.  Different choices for noise distributions, such as heavy-tailed ones for robustness, and solution preferences, such as sparsity, discussed in Section~\ref{sec:formulation}, may be considered as well.

One direction forward for computationally improving the optimization is to replace the gradient-based iterations with Newton iterations that include the Hessian of the objective in \eqref{fullLS}, which would need to be calculated in a matrix-free way.  Numerical experiments with synthetic data generated based on social theories of organization, choice, and influence may be enlightening.


\bibliographystyle{IEEEtran}
\bibliography{IEEEabrv,socialmf,references_sasha}

\end{document}